\documentclass[preprint,12pt]{elsarticle}

\usepackage{amsmath}
\usepackage{amssymb} 
\usepackage{cases}
\usepackage{url}
\usepackage{color}

\newcommand{\x}[1]{{\bf x}_{#1}}

\newcommand{\ave}[1]{\left<{#1}\right>}

\newcommand{\eq}[1]{Eq.(\ref{#1})}

\newcommand{\fig}[1]{Fig.\ref{#1}}

\newcommand{\Fig}[1]{Figure \ref{#1}}

\begin{document}

\begin{frontmatter}



\title{Computational ghost imaging using deep learning}


\author[a]{Tomoyoshi Shimobaba\corref{cor1}}
\cortext[cor1]{Tel: +81 43 290 3361; fax: +81 43 290 3361}
\ead{shimobaba@faculty.chiba-u.jp}
\author[b]{Yutaka Endo}
\author[c]{Takashi Nishitsuji}
\author[a]{Takayuki Takahashi}
\author[a]{Yuki Nagahama}
\author[a]{Satoki Hasegawa}
\author[a]{Marie Sano}
\author[a]{Ryuji Hirayama}
\author[a]{Takashi Kakue}
\author[a]{Atsushi Shiraki}
\author[a]{Tomoyoshi Ito}

\address[a]{Chiba University, Graduate School of Engineering, 1--33 Yayoi--cho, Inage--ku, Chiba, Japan, 263--8522}
\address[b]{Kanazawa University, Institute of Science and Engineering, Kakuma-machi, Kanazawa, Ishikawa 920--1192, Japan}
\address[c]{Mitsubishi Electric Corporation, Information Technology R\&D Center, 5--1--1 Ofuna, Kamakura, Kanagawa, Japan, 247--8501}

\begin{abstract}
Computational ghost imaging (CGI) is a single-pixel imaging technique that exploits the correlation between known random patterns and the measured intensity of light  transmitted (or reflected) by an object.
Although CGI can obtain two- or three- dimensional images with a single or a few bucket detectors, the quality of the reconstructed images is reduced by noise due to the reconstruction of images from random patterns.
In this study, we improve the quality of CGI images using deep learning. A deep neural network is used to automatically learn the features of noise-contaminated CGI images. After training, the network is able to predict   low-noise images from new noise-contaminated CGI images.
\end{abstract}


\end{frontmatter}

\section{Introduction}
Computational ghost imaging (CGI) \cite{shapiro2008computational} has garnered attention in recent years as a promising single-pixel imaging method. 
In CGI, we project several known random patterns onto the object to be imaged and then use a lens to collect the light transmitted an object or reflected by an object. 
The light intensities are measured by a bucket detector, such as a photodiode. 
An image of the object is then created by calculating the correlations between the known random patterns and the measured light intensities. 
CGI can image objects even in noisy environments. 

Originally, CGI only measured the light intensity of objects, but methods have also been devised for measuring its phase \cite{shirai2011ghost, clemente2012single}. 
The acquisition time for CGI schemes is long as they require a large number of illuminating random patterns to objects.
Recently, the situation has been improved by using high-speed random pattern illumination \cite{edgar2015simultaneous, wang2017high}.
In addition, three-dimensional \cite{sun20133d} and multi-spectrum CGI \cite{welsh2013fast} have been developed.

Since random patterns are used to create the object images, the reconstructed images are contaminated by noise. 
To improve the quality of CGI images, improved correlation calculation methods have been devised, such as differential \cite{ferri2010differential} and normalized CGI \cite{sun2012normalized}.
Iterative optimization schemes based on the Gerchberg--Saxton algorithm \cite{wang2015gerchberg} as well as compressed sensing \cite{welsh2013fast,zhao2012ghost} have also been applied to CGI.

In this study, we propose an approach to improve CGI image quality by using deep learning \cite{goodfellow2016deep} and confirm our technique's effectiveness through simulations. 
Deep neural networks (DNNs) can learn features for the noisy images reconstructed by CGI schemes automatically. 
We used a dataset of 15,000 images and their CGI reconstructions to train a network. 
After training, the network could predict lower-noise images from new noisy CGI images that were not included in the training set.
In Section 2, we describe our DNN-based CGI scheme.
Section 3 presents the simulation results and demonstrates the effectiveness of the proposed method.
Finally, Section 4 presents the conclusions of this study.

\section{Proposed method}

In this section, we first outline the CGI scheme used and then we describe the architecture of the DNN. 

\subsection{Computational ghost imaging}
\begin{figure}[htb]
\centerline{
\includegraphics[width=\textwidth]{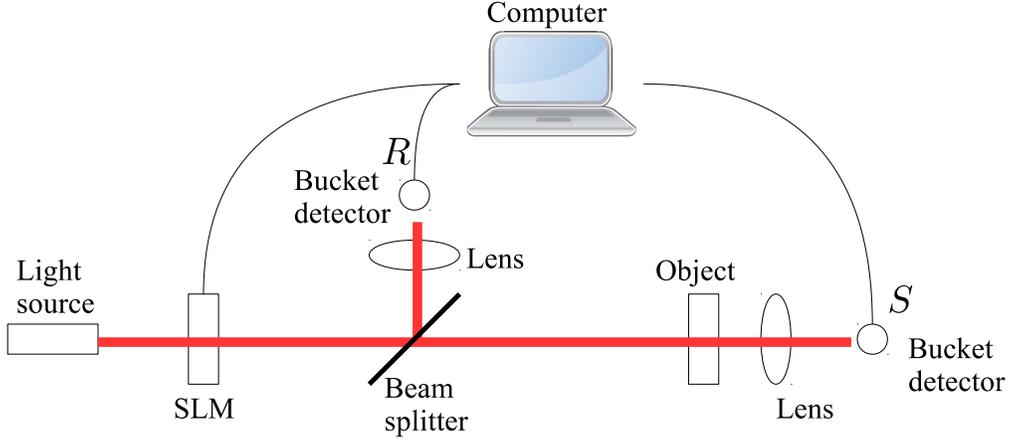}}
\caption{Optical setup for differential CGI.}
\label{fig:system}
\end{figure}

We use a differential CGI \cite{ferri2010differential} scheme because its image quality is superior to that of traditional CGI \cite{shapiro2008computational}.
The optical setup required for differential CGI is shown in \fig{fig:system}.

\begin{figure}[htb]
\centerline{
\includegraphics[width=\textwidth]{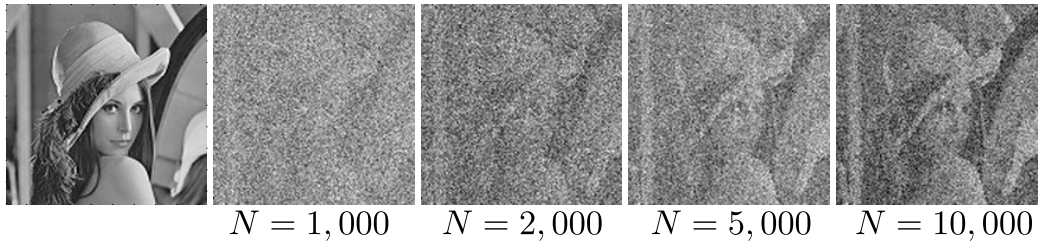}}
\caption{Example images reconstructed by differential CGI. From left to right, these are the original image that is followed by images reconstructed using $N=1,000$, $2,000$, $5,000$, and $10,000$ patterns.}
\label{fig:example}
\end{figure}

In this scheme, a sequence of random patterns is shown on a spatial light modulator (SLM).
We denote the $i$-th random pattern as $I_i(x,y)$.
The light transmitted by the SLM is divided into two beams by a beam splitter.
One beam then irradiated the object to be imaged, and the light transmitted by the object is collected by a lens, and its intensity $S_i$ is  measured by a bucket detector for each $I_i(x,y)$.
The other beam is immediately focused by a lens, and its intensity $R_i$  is measured by another bucket detector for each $I_i(x,y)$.
The final image $O(x,y)$ that is reconstructed by differential CGI is then  calculated as follows:
\begin{equation}
O(x,y)=\ave{O_i(x,y)}_N,
\label{eqn:dcgi1}
\end{equation}
where $\ave{a_i}_N=\frac{1}{N} \sum_i^N a_i$ denotes the ensemble average over all $N$ random patterns.
The $O_i(x,y)$ are calculated as follows:
\begin{equation}
O_i(x,y)=\left(\frac{S_i}{R_i} - \frac{\ave{S_i}_N}{\ave{R_i}_N} \right) 
\left( I_i(x,y) - \ave{ I_i(x,y)}_N \right).
\label{eqn:dcgi2}
\end{equation}
As can be seen from \eq{eqn:dcgi2}, the reconstructed image is expressed as a superposition of the random patterns; thus, the resulting image is noisy.
\Fig{fig:example} shows a series of example images that are reconstructed  by differential CGI. 
The images are arranged from left to right in such a manner that the original image is followed by images that are reconstructed using $N=1,000$, $2,000$, $5,000$, and $10,000$ patterns.
As the number of random pattern $N$ increases, the image quality gradually improves. However, it increases the processing and measurement time of differential CGI.

\subsection{Improving image quality using a deep neural network}
\begin{figure}[htb]
\centerline{
\includegraphics[width=\textwidth]{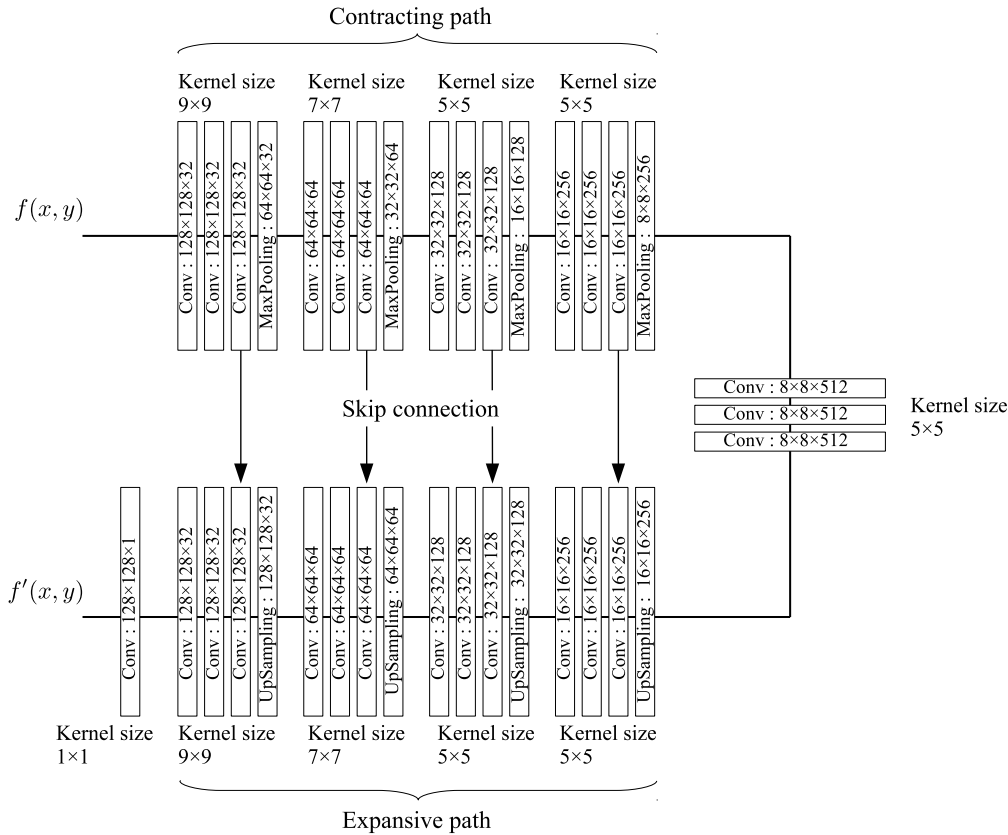}}
\caption{Our network structure \cite{ronneberger2015u}.
This network was originally used for image segmentation, but, it can be also used for image restoration \cite{jin2017deep}. }
\label{fig:network}
\end{figure}

In this study, we use a DNN to improve the quality of CGI images.
\Fig{fig:network} shows the proposed network structure which is called U-Net \cite{ronneberger2015u}. 
This network was originally used for image segmentation, but it can also be used for image restoration \cite{jin2017deep}.

The network consists of the following two paths: a constructing path and expansive path.
These paths include convolution,  max-pooling, and up-sampling layers denoted as  ``Conv'',  ``MaxPooling'' and ``UpSampling'', respectively.
The convolution layers generate feature maps for the input images using convolution operations,  which are frequently used in image processing.
For example, the first convolution layer is denoted as ``128 $\times$ 128 $\times$ 32,''which means the that it outputs 32 feature map that have an output size of 128 $\times$ 128 pixels each.
The first three convolution layers have convolution filters with a kernel size of $9 \times 9$, and the kernel weights are learned from the training dataset.  
Generally, convolution reduces the number of pixels in the output (feature map). To avoid reducing the number of pixels, we use zero padding in each convolution layer.

The output of each convolution layer except the last uses a ReLU activation function. 
The last convolution layer, which has a kernel size of $1 \times 1$, outputs a predicted image that is the same size as the input image and uses a sigmoid activation function.

The max-pooling layers down-sample the input data to reduce the influence of changes in position and size.
We used a down-sample rate of $2 \times 2$; this means that for input data of size $M \times M$, the output size is $M/2 \times M/2$.
The up-sampling layers then up-sample the input data again at a rate of $2 \times 2$; this means that for input data of size $M/2 \times M/2$, the output size is $M \times M$.

Although the max-pooling layers are important for robustness against changes in position and size in the input images, in the deeper max-pooling layers much of the input data resolution has been lost, this has resulted in them behaving like low-pass filters.
To address this drawback, skip connections have been added to the network to forward the feature maps generated by the contracting path directly to the expansive path. 

To optimize the kernel weights and other network parameters, the network is trained by minimizing the mean squared error (MSE) between the noisy images $f'(x,y)$ that are reconstructed by differential CGI and the original images $f(x,y)$.
The reconstructed images are calculated using \eq{eqn:dcgi1}.
We used  Adam optimizer \cite{kingma2014adam} to minimize the MSE using stochastic gradient descent (SGD). 
In SGD, part of datasets is randomly selected.
The size $B$ of the partial dataset is referred to as the batch size, which 
was 50 in this study. 
The number of epochs, i.e., the number of iterations used to optimize the network parameters, was 3.
In addition, we used the Dropout technique \cite{srivastava2014dropout} to prevent over-fitting in the network. 
Dropout randomly disables $d$ percent of the units in a layer during the training process, and we used $d=80\%$.

\section{Results}
To train the network, we needed to prepare a large dataset comprising pairs of original and reconstructed images.
Here, we used the Caltech-256 \cite{griffin2007caltech} dataset, which includes $\sim30,000$ general images with different resolutions.
In this study, the objects measured by differential CGI are $128 \times 128$ pixels in size; therefore, we randomly selected 15,000 images from Caltech-256 and resized them to $128 \times 128$ pixels.
These images were then reconstructed by differential CGI, using \eq{eqn:dcgi1} with 5,000 random $128 \times 128$-pixel patterns.
The reconstructed images were generated by incoherent light simulation using our numerical optics library \cite{shimobaba2012computational}.

\begin{figure}[htb]
\centerline{
\includegraphics[width=\textwidth]{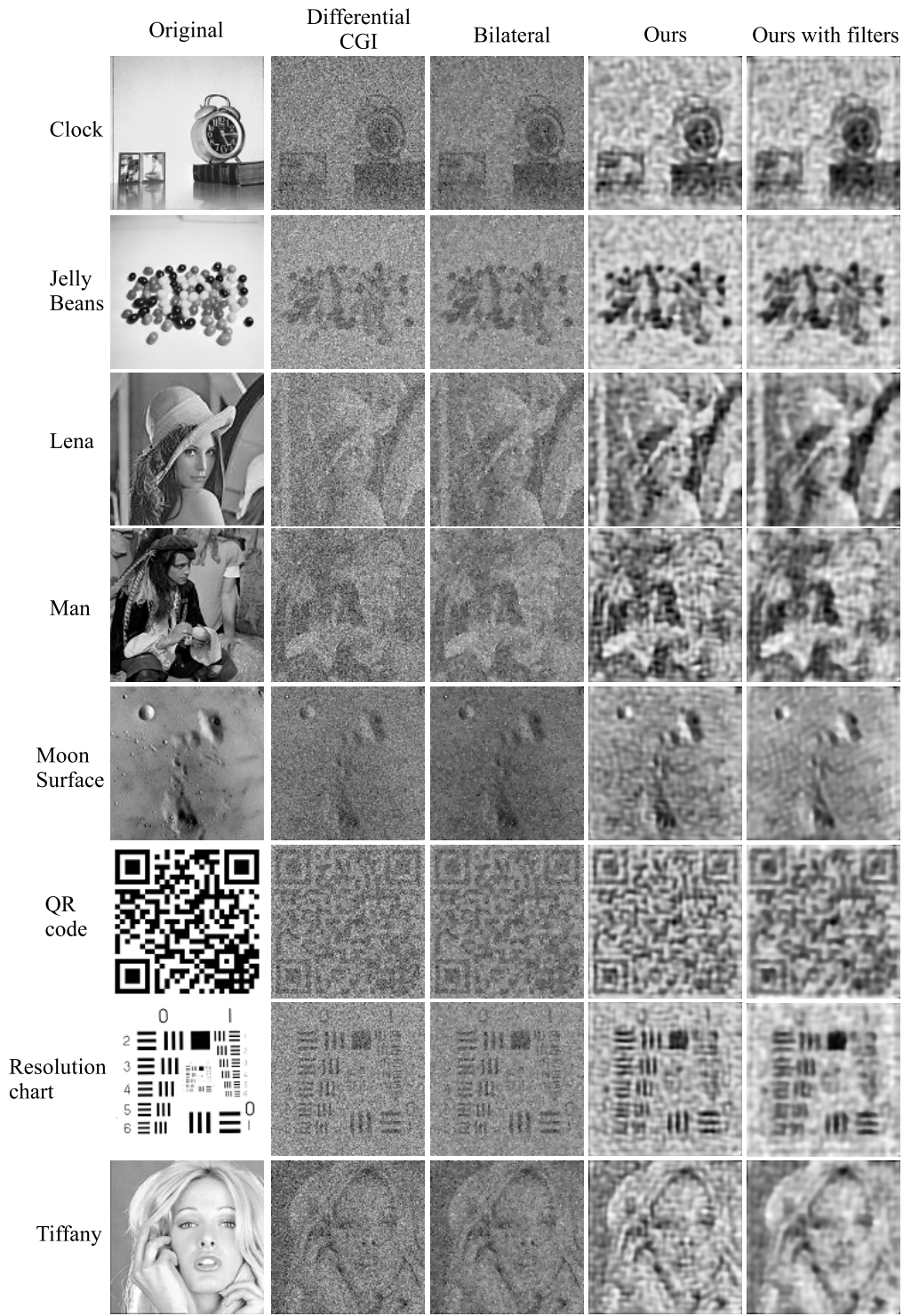}}
\caption{Comparison of the proposed methods and other methods for eight new images.}
\label{fig:result}
\end{figure}

\Fig{fig:result} shows example images that are produced using the proposed method. that is based on eight original images that were not included in the training dataset.
We compared the proposed method with a bilateral filter \cite{tomasi1998bilateral} to demonstrate its effectiveness.
Bilateral filters are a well-known type of noise-reduction filter that preserves edges.

The first column shows the eight original images, while the second shows the images reconstructed by differential CGI, the third shows the results obtained by the bilateral filter, and the fourth shows results produced by the proposed method.
Subjectively, the images obtained by the proposed method show improved noise and contrast compared with the images from the other methods.

\begin{figure}[htb]
\centerline{
\includegraphics[width=\textwidth]{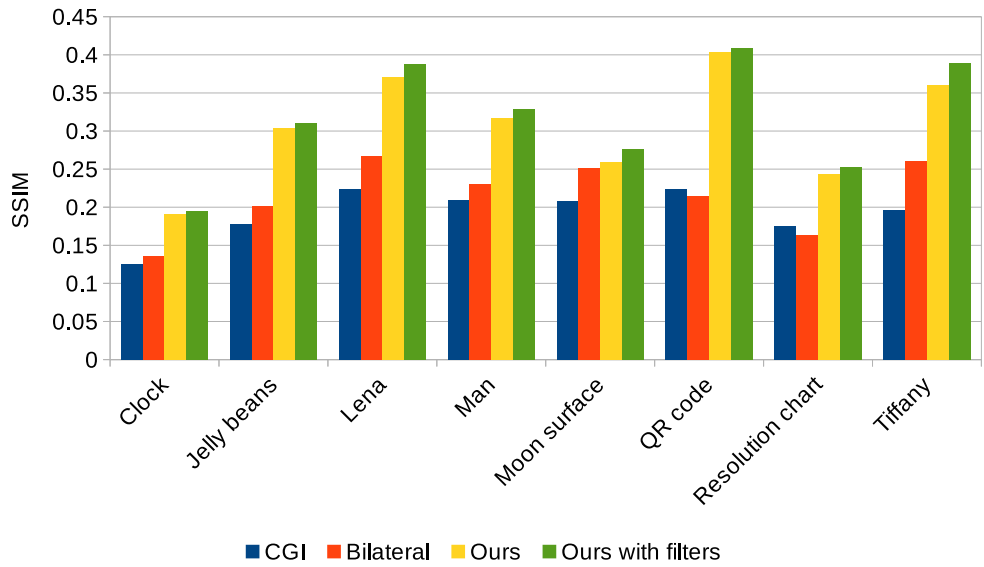}}
\caption{Comparison of the SSIM values for the proposed methods and the other methods for the eight new images.}
\label{fig:ssim}
\end{figure}

The image quality improvement achieved by the proposed method was evaluated in terms of the structural similarity (SSIM) index \cite{wang2004image}.
\Fig{fig:ssim} compares the SSIM values for the proposed method with those for the other methods.
SSIM values can evaluate image quality more accurately than the peak-signal-noise ratio (PSNR), and larger SSIMs indicate better image quality.
Overall, the SSIMs for the proposed method were better than those for the other methods for all the original images.
 
\begin{figure}[htb]
\centerline{
\includegraphics[width=\textwidth]{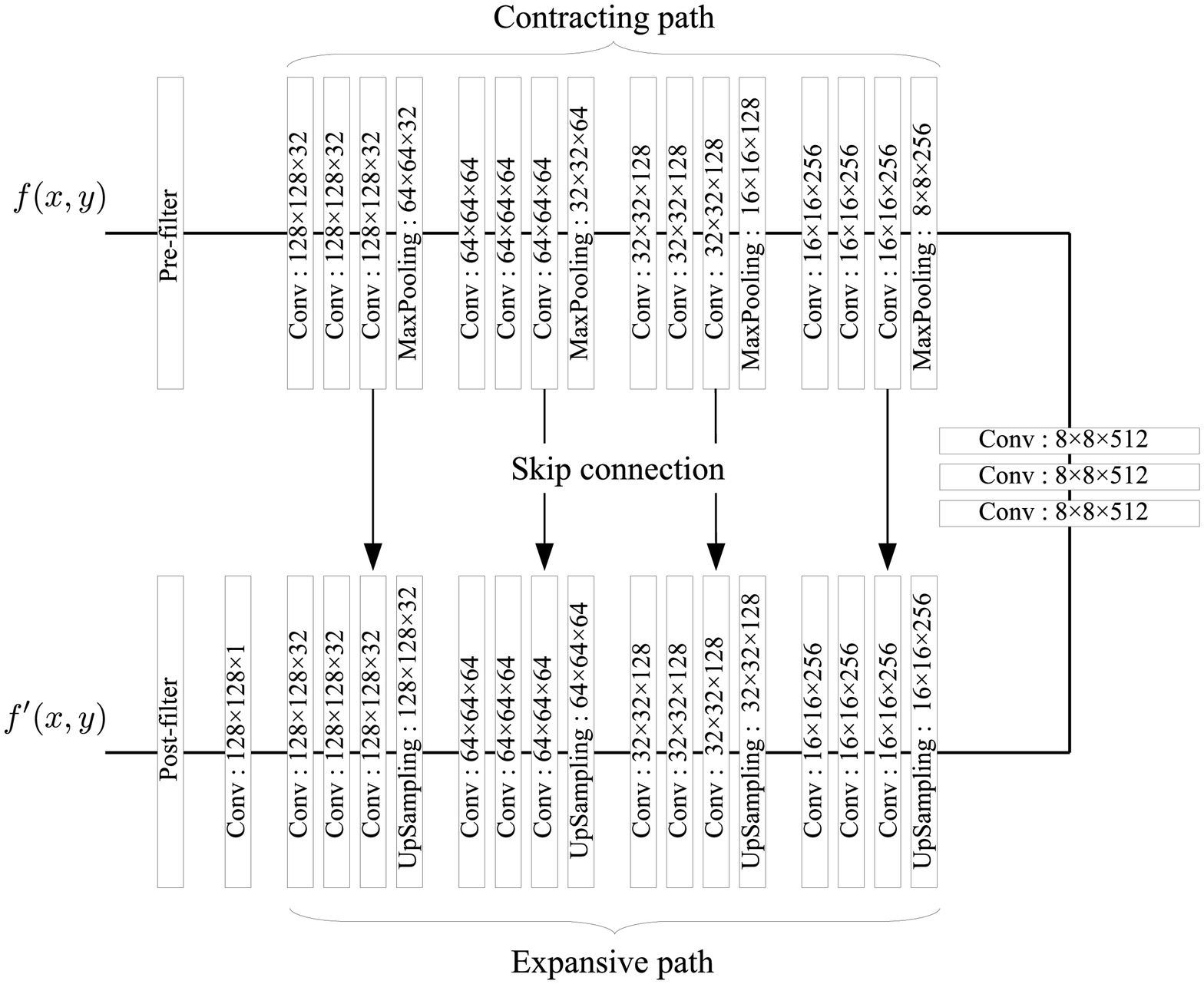}}
\caption{DNN with additional pre- and post-processing filters.}
\label{fig:network2}
\end{figure}

To improve the proposed method further, we also added pre- and post- processing filters to the DNN as shown in \fig{fig:network2}, using a bilateral filter in both cases. 
The bilateral filters reduced the noise in the training dataset, thus allowing the DNN to learn the filtered dataset.
The reconstructed images and SSIM values for the proposed method with filters are also shown in Figs. \ref{fig:result} and \ref{fig:ssim} , and these results show a small additional improvement while using the filters.
For example, the filters increases the SSIM for the ``Tiffany'' image from 0.36 to 0.39.
Table \ref{tbl:ave_ssim} shows the average SSIM values for the different methods that are averaged over the eight images. 
This again shows that our proposed methods can produce better results than the other methods.

\begin{table}[]
\centering
\caption{Average SSIMs for all methods.}
\label{tbl:ave_ssim}
\begin{tabular}{|l|l|l|l|l|}
\hline
              & CGI  & Bilateral & Ours & Ours with filters \\ \hline
Averaged SSIM & 0.19 & 0.22      & 0.31 & 0.32              \\ \hline
\end{tabular}
\end{table}

\section{Conclusions}
In this study, we have proposed using a DNN to improve the quality of images produced with CGI and have presented results from simulations where a DNN was trained using a dataset of 15,000 images. We compared the images reconstructed by the proposed method with those obtained by differential CGI and bilateral denoising. While testing with eight images that were not included in the training dataset, the average SSIM of the proposed method was over 0.3 compared with only around 0.2 for the differential CGI and bilateral denoising methods. In our next study, we will improve the structure of the DNN to further improve the image quality.

\section*{Acknowledgments}
This work was partially supported by JSPS KAKENHI, Grant Number 16K00151.

\bibliographystyle{unsrt}
\bibliography{reference.bib}

\end{document}